\documentclass{article}
\PassOptionsToPackage{numbers,sort&compress}{natbib}
\usepackage[preprint]{neurips_2026}
\usepackage[utf8]{inputenc}
\usepackage[T1]{fontenc}
\usepackage{amsmath}
\usepackage{graphicx}
\usepackage{booktabs}
\usepackage{tabularx}
\usepackage{xcolor}
\usepackage{microtype}
\usepackage{enumitem}
\usepackage{float}
\usepackage[font=small,labelfont=bf]{caption}
\usepackage[hidelinks]{hyperref}
\usepackage{url}

\newcommand{\SAN}{\text{SAN}}
\newcommand{\SANp}{\ensuremath{\text{SAN}^{+}}}
\newcommand{\SANr}{\ensuremath{\text{SAN}^{\mathrm{ref}}}}
\newcommand{\TIS}{\text{TIS}}
\newcommand{\PTH}{\text{PTH}}
\newcommand{\sg}{\operatorname{sg}}
\setlist{itemsep=0.15em,topsep=0.3em}

\title{Probe the Harness: Setup Checks for\\ Stale-Data RL Comparisons in Language Models}
\author{Taiheng Pan\\
The University of Melbourne\\
\texttt{taiheng.pan@student.unimelb.edu.au}}

\begin{document}
\maketitle

\begin{abstract}
Methods for training language models on stale samples are judged by comparisons against importance-corrected baselines. We show that details of the experimental harness can reverse the observed ranking of methods, and we introduce \PTH{} (Probe The Harness), a set of checks that makes the harness visible. Our case is a comparison between \SAN{}, a behaviour-free method, and truncated importance sampling (\TIS{}) on verl and in a single-GPU trainer, in which \SAN{} first finished ahead in both stacks. Four details of the harness changed this comparison: the PPO ratio was taken against the learner's own recomputed probabilities, the data seed did not reach the \TIS{} arm, the replay queue reused its first batch for 33 updates, and two loss normalisers differed from their description. In each case the logged quantity looked consistent with a working setup, while the quantity that defines the comparison went unchecked. With the harness checked, \TIS{} matches \SAN{} on verl, and in the trainer \TIS{} learns steadily while \SAN{} keeps a margin. We contribute the signature of each detail and its effect on the comparison, reference results for \TIS{} and uncorrected GRPO under sampler lag, and the \PTH{} checklist.
\end{abstract}

\section{Introduction}

\begin{figure}[t]
\centering
\includegraphics[width=\linewidth]{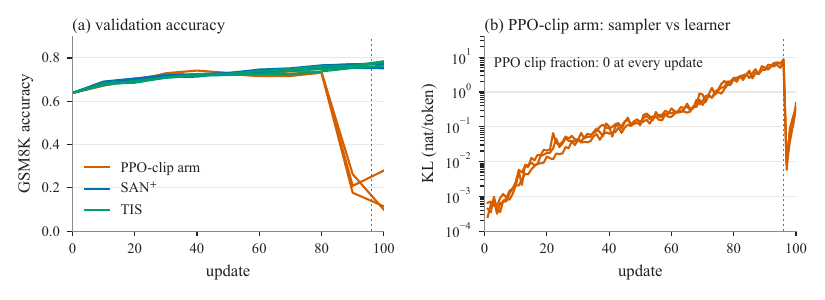}
\caption{\textbf{The PPO-clip arm trains without an active clip.} verl, Qwen2.5-Math-1.5B, GSM8K, refresh interval 96. (a) Greedy validation accuracy, one line per data seed; the dotted line marks the refresh. The PPO-clip arm (orange) drops after update 80 in every seed; \SANp{} (blue) and \TIS{} (green) end level. (b) For the PPO-clip runs, the per-token KL between sampler and learner grows by four orders of magnitude before the refresh, while the clip fraction stays at zero on every update.}
\label{fig:verl}
\end{figure}

With the sampler refreshed every 96 updates, the PPO-clip arm of our verl runs peaked at 73 to 74\% accuracy on GSM8K and then fell below 30\% in all three data seeds, while \SANp{}, a variant of the behaviour-free method we were testing, held at 75 to 77\% (Figure~\ref{fig:verl}a). In the same PPO-clip runs, the clip fraction was exactly zero on every update (Figure~\ref{fig:verl}b). That zero was the clue: the ratio was taken against the learner's own recomputed probabilities, so the clip never acted and the arm trained as GRPO without off-policy correction. With truncated importance sampling in its place, the drop disappears.

Methods for stale-data reinforcement learning are evaluated against importance-corrected baselines \citep{yao2025mismatch,zheng2026m2po,song2026collapse}, and every such comparison runs inside a harness: the code that delays the sampler, assembles each arm's configuration, selects the data for each update and computes each loss. We show that details of this harness can reverse the observed ranking of methods, through quantities that standard logs leave out. Checking those quantities is what we call \PTH{}, Probe The Harness.

Our case is \SAN{} (self-anchored normalisation), a policy-gradient method that down-weights any token whose surprisal under the learner exceeds its sequence's mean surprisal, so it needs no probabilities from the sampler. We compared it with truncated importance sampling (\TIS{}) and PPO clipping on verl \citep{sheng2024verl} and in a single-GPU trainer built to record per-token gradients. \SAN{} finished ahead in both stacks and in every variant, and it stayed stable at four model sizes. A line-by-line review of launch scripts and logs then located four details of the harness, one in each of its layers, that changed the comparison (Table~\ref{tab:layers}).

The four details share one pattern. In each layer we logged a quantity that looked consistent with a working setup, and the quantity that defines the comparison sat one step away. The clip fraction was logged; the policy in the ratio was not. Each run's label and result were logged; the seed it received was not. The data age of each update was logged; the identity of its batch was not. Each loss variant trained and logged its statistics; its agreement with the written equation was not tested. Three of these details changed the measured comparison, and the fourth changed what the comparison means.

\begin{table}[t]
\centering
\small
\caption{\textbf{Four layers of a stale-data harness.} In each layer the logged quantity looked consistent with a working setup, and the defining quantity determined the comparison. Each row has a section and a \PTH{} check.}
\label{tab:layers}
\begin{tabularx}{\linewidth}{>{\raggedright\arraybackslash}X >{\raggedright\arraybackslash}X >{\raggedright\arraybackslash}X cc}
\toprule
Layer & Logged quantity & Defining quantity & Sec. & Check \\
\midrule
Policy in the PPO ratio & Clip fraction, zero throughout & Ratio to the sampler's recorded probability & \ref{sec:ratio} & 1 \\
Configuration of each arm & Run labels, three distinct results & Data seed each arm received & \ref{sec:seed} & 2 \\
Data for each update & Data age, as designed & Identity of the training batch & \ref{sec:queue} & 3 \\
Implemented loss & Short run trains and logs & Loss equals its written equation & \ref{sec:loss} & 4 \\
\bottomrule
\end{tabularx}
\end{table}

With each defining quantity checked, the comparison reads differently. \TIS{} ends at 76 to 79\% at refresh interval 96, level with \SANp{} (Figure~\ref{fig:verl}a, green). The separation at refresh interval 64 closes once \TIS{} runs with its intended seeds. In the trainer, \TIS{} learns steadily once each update draws a new batch (Figure~\ref{fig:replay}b), and \SAN{} keeps a margin of 7 to 10 points (Section~\ref{sec:revised}).

The verl documentation describes the ratio taken against a recomputed policy as a common implementation pattern \citep{verl2025rollcorr,liu2025speedkills}, and the importance-corrected baselines in the published stale-data studies we checked take the ratio against the sampler \citep{zheng2026m2po,song2026collapse,mugrpo2026,liu2026asympo,a3po2025}. This report extends that guidance from one layer to four and contributes:
\begin{itemize}[leftmargin=1.5em]
\item the signature of each detail and its effect on a complete comparison (Section~\ref{sec:details});
\item reference results for correctly configured \TIS{} and for uncorrected GRPO at refresh intervals 64 and 96 on verl (Section~\ref{sec:revised});
\item \PTH{}, seven checks built on one principle: measure the quantity that defines the comparison (Section~\ref{sec:pth}).
\end{itemize}

\section{The harness and the method under test}
\label{sec:setup}

\paragraph{verl.} We used verl \citep{sheng2024verl} with vLLM~0.11 \citep{kwon2023vllm} and GRPO advantages \citep{shao2024deepseekmath} on Qwen2.5-Math-1.5B \citep{yang2024qwen25math} and GSM8K \citep{cobbe2021gsm8k}: 16 prompts and 8 responses per update, 512 prompt and 512 response tokens, a rule-based reward on the last number in the response, AdamW at learning rate $2{\times}10^{-6}$ unless stated, a KL penalty to the initial model with coefficient $10^{-3}$, and one optimiser step per batch. To lag the sampler, the trainer transfers the learner's weights into the inference engine only on every $N$-th update, which we call the refresh interval; between transfers the sampler is frozen, and the first $N$ updates sample from the initial model. The sampler draws at temperature 1 and records the probability of each sampled token. Validation runs greedy on the 1319 test problems every 10 updates on freshly transferred weights; the stashed sampler weights are reloaded for the next training update, so validation leaves the sampler's lag unchanged. Runs last 100 updates unless stated. The PPO-clip arm uses verl's PPO loss with clip range 0.2 \citep{schulman2017ppo}. The \TIS{} arm multiplies that loss by the detached weight $\min(\pi_{\mathrm{old}}(y_t)/q(y_t), 2)$, where $q$ is the probability the sampler recorded \citep{yao2025mismatch}. Data seeds are verl's default order and seeds 43 and 44.

\paragraph{The single-GPU trainer.} The trainer generates with Hugging Face models and updates all parameters on one GPU, which lets it record the parameter-space gradient of individual tokens. Staleness comes from a replay queue: each update generates a fresh batch with the current policy and trains on a batch generated $k$ updates earlier, which we call the replay age. The configuration in this report is Qwen2.5-Math-1.5B on GSM8K, 6 prompts and 8 responses per update, 512 response tokens, AdamW at $10^{-5}$, global gradient clipping at norm 1, no KL penalty and no reference model, 100 updates. The original grid reports the mean reward of freshly generated batches over the last 10 updates; the rerun in Section~\ref{sec:revised} adds greedy accuracy on a fixed random subset of 256 GSM8K test problems. In this trainer \TIS{} is REINFORCE with the detached weight $\min(\pi_\theta/q, 2)$.

\paragraph{\SAN{}.} For a response of $T$ tokens let $\ell_t = -\log \pi_\theta(y_t \mid x, y_{<t})$ be the learner's surprisal, $\hat A$ the group-normalised advantage of the response, and $\sg$ the stop-gradient. \SAN{} weights each token against its sequence's mean surprisal and normalises by a robust mean:
\begin{equation}
\bar S = \frac{1}{T}\sum_t \sg(\ell_t),\quad
w_t = \min\!\Big(1, \frac{\bar S}{\sg(\ell_t)}\Big),\quad
S_{\mathrm{rob}} = \frac{\sum_t w_t \ell_t}{\sum_t w_t},\quad
\mathcal L_{\SAN} = \hat A\,\frac{S_{\mathrm{rob}}}{\sg(S_{\mathrm{rob}})}.
\end{equation}
The weighted surprisal of every token is capped at the sequence mean, $w_t\ell_t=\min(\ell_t,\bar S)$, and no sampler probability enters. Two variants, \SANp{} and \SANr{}, anchor to a frozen reference model; Appendix~\ref{app:san} gives their losses as implemented.

\section{The initial comparison}
\label{sec:initial}

Table~\ref{tab:initial} lists the initial comparison. \SAN{} finished ahead in both stacks and in every variant, and at refresh interval 64 all nine \SAN{}-family runs finished above all six baseline runs. The trainer diagnostics suggested a mechanism for the \TIS{} drop: the effective sample size of the importance weights decreased, many token coefficients approached zero, and gradient norms then rose sharply. Section~\ref{sec:details} traces each of these results to a layer of the harness.

\begin{table}[t]
\centering
\small
\caption{\textbf{Initial comparison.} Final greedy accuracy on verl (one entry per run) and last-10-update reward in the trainer.}
\label{tab:initial}
\begin{tabular}{llcc}
\toprule
Stack & Setting & Baseline & \SAN{} family \\
\midrule
verl & refresh interval 64 & PPO clip: .743 .778 .777 & \SAN{}: .805 .806 .787 \\
 & & \TIS{}: .763 .776 .770 & \SANp{}: .791 .792 .784 \\
 & & & \SANr{}: .788 .803 .794 \\
verl & refresh interval 96 & PPO clip: .114 .280 .099 & \SANp{}: .773 .752 .772 \\
verl & interval 64, 300 updates & PPO clip: .723 (peak .822) & \SANp{}: .869 \\
trainer & 1.5B, replay age 32 & \TIS{}: .002 .000 .002 .002 & \SAN{}: .858 .731 .781 \\
\bottomrule
\end{tabular}
\end{table}

\section{Four layers of the harness}
\label{sec:details}

Each subsection follows one row of Table~\ref{tab:layers}: what the harness did, how it shaped the comparison, and what to check.

\subsection{Policy in the PPO ratio}
\label{sec:ratio}

verl offers two ways to obtain the ``old'' policy in the PPO ratio $r_t = \pi_\theta(y_t)/\pi_{\mathrm{old}}(y_t)$: the probabilities the sampler recorded, or probabilities recomputed by the learner at the start of the update. We used the second. With one optimiser step per batch, that step is taken at the parameters that computed $\pi_{\mathrm{old}}$, so $r_t = 1$ for every token and the clip stays inactive whatever the sampler's lag. The logs confirm it: across all 21 baseline runs with a lagged sampler, the clip fraction and the PPO KL are zero on every update, while at refresh interval 96 the KL from sampler to learner, which verl estimates as the mean of $\log q - \log \pi_{\mathrm{old}}$ over sampled tokens, reaches 7 to 9 nats per token before the refresh (Figure~\ref{fig:verl}b). The PPO-clip arm is therefore uncorrected GRPO, and we use that name from here on. \TIS{} takes its weight from the sampler's probabilities, and at refresh interval 96 it ends at .764, .781 and .785, alongside \SANp{} at .773, .752 and .772.

\emph{Check.} Log the distribution of $\pi_\theta/q$ against the sampler's recorded probability $q$. A clip fraction of exactly zero next to a sampler-learner KL that grows by orders of magnitude points to this configuration, and the source of $\pi_{\mathrm{old}}$ in the code settles it. The verl documentation also describes this configuration \citep{verl2025rollcorr,liu2025speedkills}. \PTH{} check 1.

\subsection{Configuration of each arm}
\label{sec:seed}

The verl launch script assembled its extra arguments in a variable, appended the data seed, and, when importance sampling was enabled, assigned a new string to the same variable. The seed therefore reached every arm except \TIS{}. The \SAN{} and PPO-clip arms at refresh interval 64 ran three data orders, and the three \TIS{} runs ran one data order three times, differing through sampling randomness (.763, .776, .770). With seeds 43 and 44 restored, \TIS{} ends at .785 and .808 (Table~\ref{tab:verl}), and the separation closes: paired by seed, the \SAN{} family leads by 2.6 to 4.2 points on the default order and by 0.8 to 2.1 on seed 43, \TIS{} leads by 1.4 to 2.4 on seed 44, and the mean difference is 0.9 points.

\emph{Check.} Print the resolved configuration of every run and compare the ordered prompt IDs of each arm's first batches. Runs labelled with different seeds share that sequence when the seed has not reached them. \PTH{} check 2.

\subsection{Data for each update}
\label{sec:queue}

\begin{figure}[t]
\centering
\includegraphics[width=\linewidth]{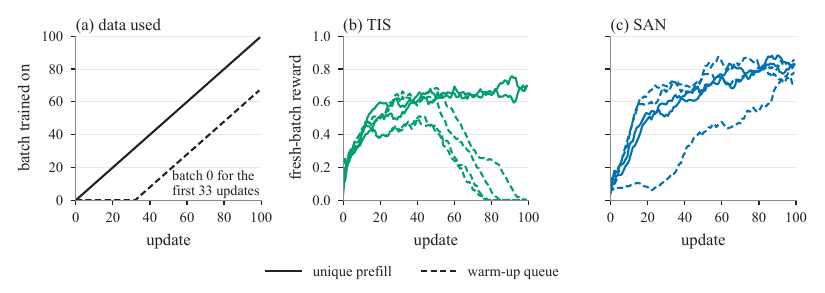}
\caption{\textbf{The replay warm-up accounts for the \TIS{} drop.} Single-GPU trainer, Qwen2.5-Math-1.5B, GSM8K, replay age 32. (a) Index of the batch each update trains on. Both queues give every update the same data age; the warm-up queue trains on batch 0 for the first 33 updates. (b, c) Reward on freshly generated batches, 10-update running mean, under the unique-prefill queue (solid, two runs per method) and the warm-up queue (dashed, four \TIS{} and three \SAN{} runs). Under the warm-up queue every \TIS{} run falls to zero after update 50; under unique prefill both keep learning. \SAN{} is similar under both queues.}
\label{fig:replay}
\end{figure}

The trainer's replay queue appended each new batch and trained on the oldest one once it held $k{+}1$ batches. Until then it trained on the oldest batch available, batch 0, so the first $k{+}1$ updates all used the same 48 responses. The data age followed the intended schedule from 0 to $k$ throughout. At replay age 32, 100 updates used 68 distinct batches (Figure~\ref{fig:replay}a), and every staleness result in the original grid used this queue.

A unique-prefill queue generates $k{+}1$ batches from the initial policy before training and then gives every update a new batch, with the same data-age schedule. With the same model, learning rate, batch size and seeds, \TIS{} learns steadily under this queue: its last-10 reward is .70 in both seeds, compared with .00 in the four warm-up runs, and its held-out accuracy at update 100 is 68.4 and 67.2\% (Figure~\ref{fig:replay}b). \SAN{} performs similarly under both queues, at .86, .73 and .78 with the warm-up and .83 and .83 with unique prefill. Repeated training on one batch acts on the method that weights each token by its ratio to the sampler, and the method that does not read the sampler stays steady.

\emph{Check.} Log the identity of each training batch and count distinct batches. When the design calls for a new batch at every update, fewer distinct batches than updates identifies the warm-up, even with a data age that matches the design. The drop appeared after update 50, well after the warm-up ended, which makes batch identity the key log. \PTH{} check 3.

\subsection{Implemented loss}
\label{sec:loss}

The draft described \SANp{} and \SANr{} as \SAN{} with a different anchor in $w_t$ and all other components unchanged, which gives the loss $\hat A\, S_{\mathrm{rob}}/\sg(S_{\mathrm{rob}})$ for both. The code divides \SANp{}'s weighted surprisal by the total anchor mass and \SANr{}'s robust mean by the reference model's mean surprisal (Appendix~\ref{app:san}), so the anchor comparison varied the anchor and the normaliser together. This layer leaves the numbers unchanged and sets their interpretation: differences between the variants belong to both components jointly.

\emph{Check.} Evaluate each implemented loss and its gradient on a fixed input and compare them with the written equation. A short training run confirms that a variant trains; this test confirms that it is the variant described. \PTH{} check 4.

\section{Results with the harness checked}
\label{sec:revised}

\paragraph{Reference results on verl.} Table~\ref{tab:verl} gives every comparison with the harness checked, paired by data seed. Correctly configured \TIS{} is stable at refresh intervals 64 and 96, ending between .763 and .808, and uncorrected GRPO drops in every seed at interval 96. At learning rate $2{\times}10^{-6}$ the \SAN{} family and \TIS{} are level at both intervals; the paired differences for \SANp{} at interval 96 are $+0.9$, $-2.9$ and $-1.3$ points. At $4{\times}10^{-6}$, in one seed, \TIS{} ends at .837 and \SANp{} at .733.

\begin{table}[b]
\centering
\small
\caption{\textbf{verl with the harness checked.} Final greedy GSM8K accuracy at update 100, paired by data seed (default order, 43, 44). At refresh interval 64, the \TIS{} default-order entry is the original first run and the seed-43 and seed-44 entries are reruns with those seeds. Blank cells were not run.}
\label{tab:verl}
\begin{tabular}{lccccccc}
\toprule
 & \multicolumn{3}{c}{interval 64, lr $2{\times}10^{-6}$} & \multicolumn{3}{c}{interval 96, lr $2{\times}10^{-6}$} & interval 96, lr $4{\times}10^{-6}$ \\
\cmidrule(lr){2-4}\cmidrule(lr){5-7}\cmidrule(lr){8-8}
 & default & 43 & 44 & default & 43 & 44 & default \\
\midrule
Uncorrected GRPO & .743 & .778 & .777 & .114 & .280 & .099 & \\
\TIS{} & .763 & .785 & .808 & .764 & .781 & .785 & .837 \\
\SAN{} & .805 & .806 & .787 & .738 & & & \\
\SANp{} & .791 & .792 & .784 & .773 & .752 & .772 & .733 \\
\SANr{} & .788 & .803 & .794 & & & & \\
\bottomrule
\end{tabular}
\end{table}

\paragraph{The trainer.} With the unique-prefill queue, \TIS{} reaches 68.4 and 67.2\% held-out accuracy and \SAN{} 75.4 and 77.0\%. The token weights account for most of the gain over plain sequence normalisation: the weights $w_t$ alone give 77.3 and 78.1\%, against 71.5 and 70.7\% for plain normalisation (Appendix~\ref{app:san}, Table~\ref{tab:trainer}).

\paragraph{Two stacks, two regimes.} The trainer and verl differ in five respects: no KL penalty against $10^{-3}$, learning rate $10^{-5}$ against $2{\times}10^{-6}$, a constant data age after warm-up against a sawtooth from 0 to $N{-}1$, 48 responses per update against 128, and a sampler that shares the learner's numerics against a separate inference engine. The margin appears in the first regime, and locating its source calls for a factorial comparison over these five respects.

\section{\PTH{}: probe the harness}
\label{sec:pth}

\PTH{} rests on one principle: for every layer of the harness, measure the quantity that defines the comparison, alongside the quantities that are convenient to log. Checks 1 to 4 apply it to the four layers of Table~\ref{tab:layers}; checks 5 to 7 apply it to the comparison as a whole.
\begin{enumerate}[leftmargin=1.5em]
\item \textbf{Log the ratio against the policy that generated the data.} For every arm, record the clip fraction and the distribution of $\pi_\theta/q$ with $q$ the sampler's recorded probability. Under a lagged sampler, a clip fraction of exactly zero is the signal to inspect where $\pi_{\mathrm{old}}$ comes from.
\item \textbf{Check the resolved configuration per arm.} Print the full configuration each run used and confirm, per arm, that the fields the comparison varies differ and the fields it holds fixed, data seed included, match. The ordered prompt IDs of the first batches confirm the data order directly.
\item \textbf{Log the identity of the training batch alongside its age.} In a replay simulator, count the distinct batches used and confirm that the count matches the design.
\item \textbf{Test every loss against its equation.} For each variant, evaluate the implemented loss and its gradient on a fixed input and compare them with the written formula.
\item \textbf{Map each baseline's stable range.} Before comparing, find where the baseline degrades under its intended configuration. In every revised setting of this report, \TIS{} stays stable over all 100 updates.
\item \textbf{Report paired differences per seed.} Pooled rankings such as nine above six mix seed effects with method effects; per-seed pairs separate them.
\item \textbf{Review code and logs against each claim.} Reading launch scripts and logs line by line against the sentences of the draft located all four layers in this report.
\end{enumerate}

\section{Related work}

Deep RL results depend on implementation details and seeds \citep{henderson2018matters,engstrom2020implementation}, and reasoning benchmarks for language models on decoding settings, seeds and hardware \citep{hochlehnert2025sober}. \PTH{} brings the same discipline to the part of the pipeline that simulates staleness.

The gap between a lagged or mismatched sampler and the learner is corrected with truncated importance weights \citep{yao2025mismatch}, by taking the PPO ratio against the sampler \citep{zheng2026m2po,song2026collapse}, by separating the proximal policy from the behaviour policy \citep{fu2025areal,a3po2025}, or by reshaping the clipped objective \citep{mugrpo2026,zheng2026m2po}. \citet{liu2025speedkills} and the verl documentation \citep{verl2025rollcorr} describe the ratio taken against a recomputed policy as a common source of instability, which \PTH{} check 1 detects. Behaviour-free alternatives normalise by the learner's own statistics \citep{liu2026asympo} or gate on the learner's surprisal \citep{osband2026delightful}, and variance control adapts the step to the importance weights \citep{huang2026vcpo}; \SAN{} belongs to the behaviour-free family.

\section{Conclusion}

A stale-data comparison measures the harness together with the methods. In our case, four layers of the harness, namely the policy in the PPO ratio, the configuration of each arm, the data for each update and the implemented loss, changed the outcome or its interpretation, and each layer carried a defining quantity beyond what we logged. With those quantities checked, \TIS{} matches \SAN{} on verl and learns steadily in the trainer, and the margin \SAN{} keeps in the trainer is the natural target for a factorial study of the two stacks. \PTH{} turns each layer into one check, so that a comparison measures the methods it names.

\bibliographystyle{plainnat}
\bibliography{references}

\appendix
\section{SAN variants and the trainer ablation}
\label{app:san}

The two variants anchor to a frozen reference model with surprisal $\ell^{\mathrm{ref}}_t$, available in any stack that computes a KL penalty. As implemented, their losses are
\begin{align}
\SANp:&\quad a_t=\max\big(\ell^{\mathrm{ref}}_t, \bar S\big),\quad w_t=\min\!\Big(1,\frac{a_t}{\sg(\ell_t)}\Big),\quad
\mathcal L = \hat A\,\frac{\sum_t w_t\ell_t}{\sum_t a_t},\\
\SANr:&\quad w_t=\min\!\Big(1,\frac{\ell^{\mathrm{ref}}_t}{\sg(\ell_t)}\Big),\quad
\mathcal L = \hat A\,\frac{S_{\mathrm{rob}}}{\frac1T\sum_t \ell^{\mathrm{ref}}_t}.
\end{align}

Table~\ref{tab:trainer} decomposes \SAN{} in the trainer with the unique-prefill queue. Writing $S=\sum_t \ell_t$ and $D=\sum_t w_t \ell_t$, plain sequence normalisation puts the coefficient $\hat A/S$ on every $\nabla\ell_t$ and \SAN{} puts $\hat A w_t/D$. The token weights also change how often the global gradient clip activates, from 94 and 95 of 100 updates for plain normalisation to 12 and 19. Fixing the gradient norm at 1 before every Adam step narrows the margin of token weights over plain normalisation from 5.9 to 2.0 points in one seed and from 7.4 to 6.3 in the other, so the global gradient scale explains part of the margin in one seed and little of it in the other. The stabilised loss of ASymPO \citep{liu2026asympo} is close to \TIS{}.

\begin{table}[H]
\centering
\small
\caption{\textbf{Trainer, unique-prefill queue.} Replay age 32, Qwen2.5-Math-1.5B, no KL penalty. Greedy accuracy (\%) on 256 GSM8K test problems at update 100, seeds 404 and 505; 21.5\% at update 0.}
\label{tab:trainer}
\begin{tabular}{llccc}
\toprule
Method & Coefficient & 404 & 505 & Mean \\
\midrule
\TIS{} & $\hat A \min(\pi_\theta/q,2)$ & 68.4 & 67.2 & 67.8 \\
ASymPO, stabilised loss & & 69.9 & 67.2 & 68.6 \\
Plain sequence normalisation & $\hat A/S$ & 71.5 & 70.7 & 71.1 \\
Sequence gain only & $\hat A/D$ & 74.6 & 74.2 & 74.4 \\
\SAN{} & $\hat A w_t/D$ & 75.4 & 77.0 & 76.2 \\
Token weights only & $\hat A w_t/S$ & 77.3 & 78.1 & 77.7 \\
\midrule
\multicolumn{5}{l}{\emph{Gradient norm set to 1 before every Adam step}} \\
Plain sequence normalisation & $\hat A/S$ & 74.2 & 72.3 & 73.2 \\
Token weights only & $\hat A w_t/S$ & 76.2 & 78.5 & 77.3 \\
\bottomrule
\end{tabular}
\end{table}

\section{The original trainer grid at replay age 32}
\label{app:grid}

All runs below used the warm-up queue of Section~\ref{sec:queue} and are reported as recorded: mean reward of freshly generated batches over the last 10 of 100 updates, seeds in parentheses. The 3B and 7B models used 8-bit or paged Adam and shorter responses to fit one GPU, and the grid also covered Qwen2.5-0.5B at replay age 16, where \SAN{} reached .46 to .59 across five seeds. Outside the 1.5B full-parameter configuration, \TIS{} stays well above zero, and its difference from \SAN{} varies in sign across these single runs, from 19.6 points behind to 9.6 points ahead.

\begin{table}[H]
\centering
\footnotesize
\setlength{\tabcolsep}{4pt}
\begin{tabularx}{\linewidth}{l c >{\raggedright\arraybackslash}X >{\raggedright\arraybackslash}X}
\toprule
Model & lr & \TIS{} & \SAN{} \\
\midrule
Qwen2.5-Math-1.5B & $10^{-5}$ & .002 (404), .000 (505), .002 (404), .002 (505) & .858 (404), .731 (505), .781 (606) \\
\quad later code version & $10^{-5}$ & .273 (606), .108 (707), .004 (808), .150 (909) & .833 (707) \\
Qwen2.5-3B-Instruct & $5{\times}10^{-6}$ & .821 (404) & .860 (404) \\
Qwen2.5-Math-7B & $3{\times}10^{-6}$ & .569 (404) & .616 (404) \\
Llama-3.2-3B-Instruct & $5{\times}10^{-6}$ & .800 (404), .717 (505) & .771 (404) \\
Qwen2.5-Math-1.5B, LoRA 64 & $10^{-4}$ & .817 (404), .619 (505) & .721 (404), .815 (505) \\
\bottomrule
\end{tabularx}
\end{table}

\end{document}